\documentclass[11pt]{article}

\usepackage[final]{acl}

\usepackage{times}
\usepackage{latexsym}

\usepackage[T1]{fontenc}

\usepackage[utf8]{inputenc}

\usepackage{microtype}

\usepackage{inconsolata}

\usepackage{graphicx}

\usepackage{algorithm}
\usepackage{enumitem}
\usepackage{newfloat}
\usepackage{listings}
\usepackage[most]{tcolorbox}
\usepackage{xcolor}  
\usepackage{booktabs}
\usepackage{multirow}
\usepackage{amsmath, amssymb}
\usepackage{booktabs}
\usepackage[table]{xcolor}
\usepackage{tabularx}
\usepackage{pifont}
\usepackage{algorithm}
\usepackage{algpseudocode} 
\newcommand{\M}{\textsc{ScaffoldAgent}}
\title{ScaffoldAgent: Utility-Guided Dynamic Outline Optimization for Open-Ended Deep Research}

\author{
  \textbf{Zhibang Yang\textsuperscript{1,2,3,*}},
  \textbf{Xinke Jiang\textsuperscript{1,2,3,*}},
  \textbf{Yuzhen Xiao\textsuperscript{1,2,3,*}},
  \textbf{Ruizhe Zhang\textsuperscript{1,2,3,*}},
  \textbf{Yue Fang\textsuperscript{1,2,*}},
  \textbf{Xinfei Wan\textsuperscript{1}},\\
  \textbf{Zhengxing Song\textsuperscript{1}},
  \textbf{Yuxuan Liu\textsuperscript{1}},
  \textbf{Yuheng Huang\textsuperscript{4}},
  \textbf{Junfeng Zhao\textsuperscript{2,3,\ensuremath{\dagger}}},
  \textbf{Yasha Wang\textsuperscript{1,5,\ensuremath{\dagger}}},
  \textbf{Xu Chu\textsuperscript{2,3,6,\ensuremath{\dagger}}}\\
  \textsuperscript{1}National Engineering Research Center of Software Engineering, Peking University, Beijing, China\\
  \textsuperscript{2}School of Computer Science, Peking University, Beijing, China\\
  \textsuperscript{3}Key Laboratory of High Confidence Software Technologies, Ministry of Education, Beijing, China\\
  \textsuperscript{4}GRG Banking Equipment Co., Ltd., Guangzhou, China\\
  \textsuperscript{5}Peking University Information Technology Institute (Tianjin Binhai), Tianjin, China\\
  \textsuperscript{6}Center on Frontiers of Computing Studies, Peking University, Beijing, China\\ \small
  \textsuperscript{*}Equal contribution \quad
  \textsuperscript{\ensuremath{\dagger}}Corresponding authors\\
  \small{\texttt{\{yangzb, xinkejiang\}@stu.pku.edu.cn;
  \{zhaojf, wangyasha, chu\_xu\}@pku.edu.cn}}
}

\begin{document}
\maketitle 
\begin{abstract}
Open-ended deep research (OEDR) requires systems to acquire knowledge through multi-round retrieval and generate coherent long-form reports. The outline plays a central role as a structural scaffold that coordinates retrieval, evidence organization, and generation. However, existing methods either fix the outline before writing or refine it with local heuristics, leading to scaffold drift under continuous information accumulation and delayed feedback for evaluating outline modifications. We propose \textbf{ScaffoldAgent}, a utility-guided dynamic outline optimization framework for OEDR. ScaffoldAgent models outline evolution as a structured decision process with three operations: \emph{Expansion}, \emph{Contraction}, and \emph{Revision}, enabling controlled updates to the report scaffold. It further introduces a utility-guided feedback mechanism that estimates the downstream value of each outline operation from retrieval gain, structural coherence, and trial-generation quality. The resulting utility signal guides node selection, operation scheduling, and termination during inference. Experiments on DeepResearch Bench and DeepResearch Gym show that ScaffoldAgent consistently improves long-form report generation and factual grounding over existing deep research agents.
\end{abstract}

\section{Introduction}

\begin{figure}[t]
    \centering
 \includegraphics[width=0.48\textwidth]{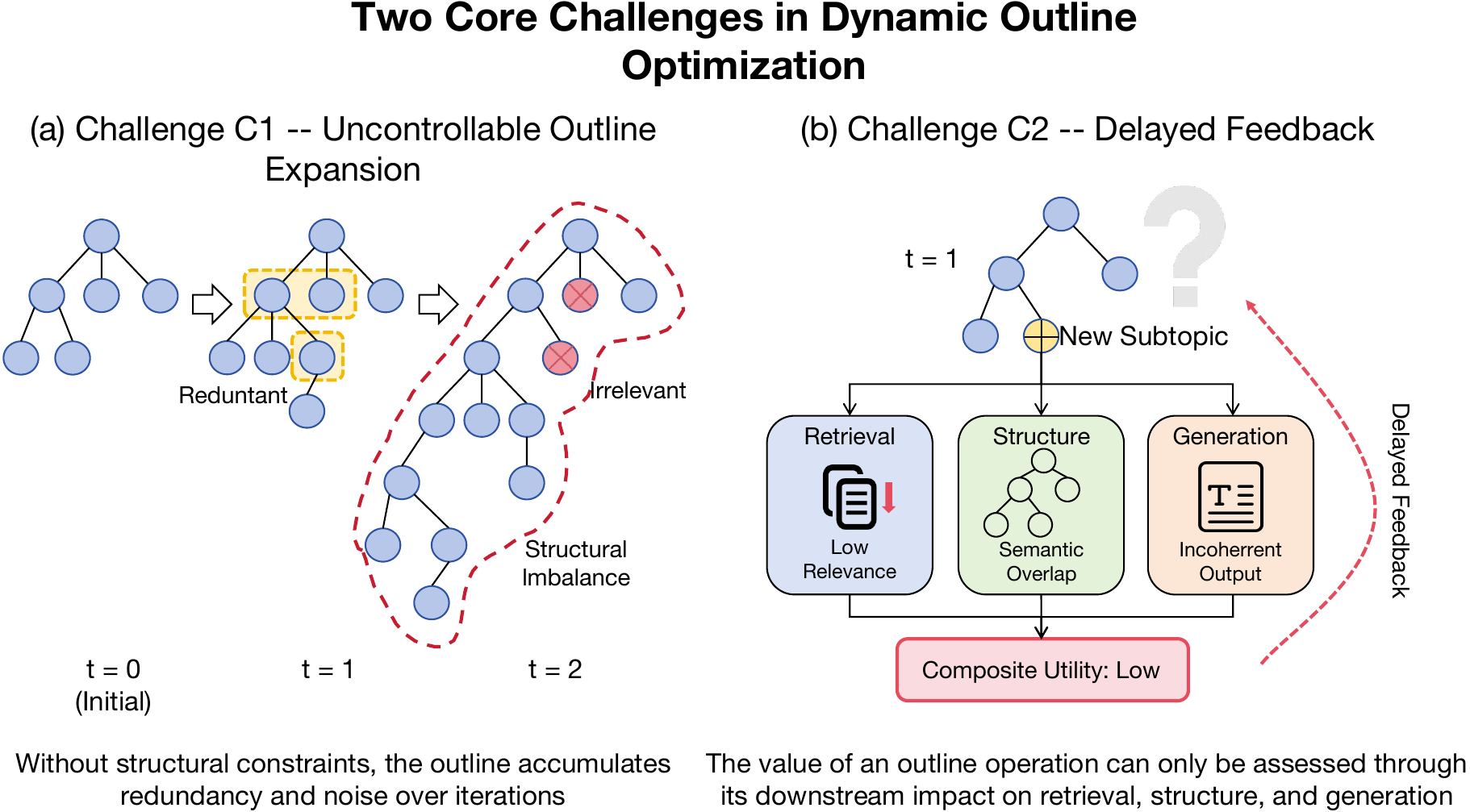}
\caption{Challenges in dynamic outline optimization. 
\textbf{C1: Scaffold Drift.} As the outline is repeatedly updated with newly retrieved evidence and subtopics, redundant branches, uneven topic granularity, and misplaced sections gradually accumulate. 
\textbf{C2: Delayed Feedback.} An outline modification yields no immediate feedback, as its impact only becomes observable through subsequent retrieval and trial generation.
}
    \label{fig:challenges}
\end{figure}

\textbf{L}arge \textbf{L}anguage \textbf{M}odels (\textbf{LLMs}) have become a cornerstone for building increasingly capable language agents~\cite{bubeck2023paper,chang2024survey}. Recent advances in reasoning-oriented models, such as DeepSeek-R1~\cite{guo2025deepseek} and the OpenAI o-series~\cite{jaech2024openai}, have substantially improved LLM performance on structured tasks such as mathematical reasoning and complex question answering~\cite{trinh2024solving,guo2026rethinking}. Beyond single-turn generation, tool-augmented LLMs, i.e., agents~\cite{li2026agentcpm}, can now interact with external environments and perform iterative planning, searching, and self-correction~\cite{jiang2024tc,yao2023react,zhang2026stackplanner,shinn2024reflexion,schick2023toolformer,liu2026proprietary,feng2026trust}. 

This agentic paradigm has shown strong potential in software engineering~\cite{jimenez2024swe} and open-world web navigation~\cite{zhou2024webarena,mialon2024gaia}. However, \textbf{open-ended tasks} that require agents to acquire knowledge through multi-round retrieval and synthesize the accumulated evidence into coherent long-form reports \textbf{remain challenging}. Such tasks, recently formalized as \textbf{Open-ended Deep Research (OEDR)}~\cite{li2025webweaver,du2025deepresearch}, demand not only effective information seeking and content generation, but also persistent \textbf{content organization} over evolving evidence.

Unlike task-oriented agents that execute a fixed plan toward a well-defined goal~\cite{jimenez2024swe,mialon2024gaia}, OEDR is organized around an open-ended objective whose \textbf{scope evolves as new evidence is discovered}. In each round, the system must retrieve new information, reconcile it with prior reasoning, and update the organization of the emerging report~\cite{li2025webweaver,li2026agentcpm,prabhakar2025enterprise}. Within this loop, the outline is not merely a pre-writing plan but an active structural scaffold: it guides what to search for, indexes evidence under appropriate sections, and constrains final report generation~\cite{shao2024assisting,xiong2025beyond,wang2025generating}. 

However, existing deep research systems still face substantial limitations. 
As a report grows in length and topical breadth, a static or poorly maintained outline may gradually \textbf{drift from the evolving evidence space}, leading to redundant coverage, missing perspectives, or incoherent section organization.
One line of work follows a plan-then-write paradigm, where the outline is fixed before report generation begins~\cite{shao2024assisting,lee2025navigating,han2025deep}, causing the planned outline to gradually diverge from the evolving information space~\cite{xiong2025beyond}. Another line of work introduces dynamic outline update mechanisms but relies predominantly on heuristic rules or local feedback~\cite{wang2025generating, shi2025scisage, chen2025surveygen, wan2025cognitive} and lacks a unified optimization objective that jointly accounts for retrieval quality, structural soundness, and generation quality~\cite{wu2025superwriter, xiong2025beyond}: for instance, updating the outline based solely on newly retrieved documents~\cite{li2025webweaver, wang2025generating}, or on signals that do not propagate beyond the current step~\cite{wan2025cognitive, chen2025surveygen}.  

This limitation gives rise to two core challenges, as illustrated in Figure~\ref{fig:challenges}. First, without explicit control over structural evolution, iterative refinement may cause \textbf{scaffold drift}: newly retrieved evidence can trigger redundant branches, uneven topic granularity, or misplaced subtopics, progressively destabilizing the global hierarchy of the outline (\textbf{C1}). Second, the usefulness of an outline modification is difficult to assess at the moment it is made, since any section modification may reveal its value only after subsequent retrieval and trial generation, resulting in \textbf{delayed feedback} for evaluating downstream utility (\textbf{C2}).
 
To address these challenges, we propose \textbf{ScaffoldAgent}, a utility-guided framework for dynamic outline optimization in open-ended deep research. Rather than treating the outline as a fixed pre-writing plan, ScaffoldAgent maintains it as an evolving structural scaffold that coordinates retrieval, evidence organization, and report generation throughout the research process. To mitigate \textbf{C1}, ScaffoldAgent formulates outline evolution as a structured decision process with three explicit operations: \emph{Expansion}, \emph{Contraction}, and \emph{Revision}. Inspired by classical belief revision theory~\cite{alchourron1985logic}, these operations enable the system to incorporate new information, merge redundant branches, and revise weakly supported sections while preserving hierarchical consistency. To address \textbf{C2}, ScaffoldAgent introduces a utility-guided feedback mechanism that estimates the usefulness of each outline modification from retrieval gain, structural coherence, and trial-generation quality. A Reporter Agent performs trial writing under the current outline, and its generation quality and citation grounding provide downstream feedback. The resulting utility estimate serves as an inference-time control signal for node selection, operation scheduling, and termination, enabling ScaffoldAgent to optimize outline evolution beyond local heuristic refinement.
Contributions are as follows:
\begin{itemize}[leftmargin=*,noitemsep,topsep=2pt]

\item We propose \textbf{ScaffoldAgent}, a utility-guided framework that treats the outline as an evolving structural scaffold and optimizes it through three explicit operations: \emph{Expansion}, \emph{Contraction}, and \emph{Revision}.

\item We introduce a utility-guided feedback mechanism that evaluates outline operations along retrieval, structural, and generation dimensions, enabling inference-time control over node selection, operation scheduling, and termination.
 
\item Experiments on DeepResearch Bench and DeepResearch Gym show that ScaffoldAgent consistently outperforms existing deep research agents, validating the effectiveness of utility-guided outline optimization for improving long-form report generation and factual grounding.
\end{itemize}


\section{Related Work}

\subsection{Open-Ended Deep Research}
Open-Ended Deep Research (OEDR) extends bounded question answering to long-horizon evidence collection and grounded report synthesis, as formalized by recent benchmarks~\cite{du2025deepresearch, coelho2025deepresearchgym}. Built upon tool-augmented agents and search-augmented reasoning ~\cite{yao2023react,trivedi2023interleaving,jin2025search, zhang2026stackplanner}, existing OEDR systems have explored several directions. STORM and TTD-DR follow plan- or refinement-oriented workflows for long-form research report generation \cite{shao2024assisting, han2025deep}. WebWeaver and AgentCPM-Report further introduce dynamic outlines to connect evidence acquisition with report writing, either through planner-writer collaboration or drafting-driven deepening \cite{li2025webweaver,li2026agentcpm}. EDR, RhinoInsight, and FS-Researcher enhance OEDR from the perspectives of multi-agent coordination, behavior/context control, and persistent external memory \cite{prabhakar2025enterprise, lei2025rhinoinsight, zhu2026fs}. While they improve retrieval, memory, agent coordination, or writing feedback, outline updates are still largely driven by fixed workflows, local heuristics, or single-stage feedback. They do \textbf{not unify retrieval gain, structural coherence, and trial-generation quality into single utility signal for node selection, operation scheduling, and termination}. 

\subsection{Report Generation}
Grounded report generation builds on retrieval-augmented and attributed generation, where external evidence improves factuality and supports citation-aware writing \cite{lewis2021retrievalaugmented, menick2022teaching, gao2023enabling,liu2025generalization,xu2025parenting,yang2026dfams,zhang2025knowpo}. For long-form report generation, a common solution is the plan-then-write paradigm, which first constructs an outline or high-level report plan and then generates content section by section to maintain global organization \cite{yao2019plan,yang2023doc,yang2022re3,bai2025longwriter, wan2025cognitive}. More adaptive methods further update outlines, memory, or task graphs during extended generation: DOME uses dynamic hierarchical outlines with memory enhancement, WriteHERE performs heterogeneous recursive planning, and survey-generation systems such as SurveyGen-I and SciSage use evolving plans, memory-guided writing, or reflection to improve cross-section coherence, coverage, and citation quality \cite{wang2025generating, xiong2025beyond, chen2025surveygen,shi2025scisage}. These methods improve structure and revision of long reports, \textbf{but mainly optimize generated text or local writing plan}. 


\section{OEDR Task Formulation}
\label{sec:problem}

Given a research question $q$ and access to a large-scale document collection $\mathcal{D}$, \textbf{Open-ended Deep Research (OEDR)} aims to generate a structurally coherent and well-cited long-form report. 
Unlike one-shot retrieval-augmented generation, OEDR requires a system to iteratively search for evidence, organize the acquired information, and synthesize a coherent final report.

We formulate OEDR as a dynamic interaction among three components: \emph{Search}, \emph{Outline}, and \emph{Report}. 
Search retrieves external evidence from the document collection; Outline organizes the accumulated evidence into a hierarchical scaffold that defines the scope and structure of the report; and Report transforms the organized evidence into coherent long-form text. 
These three components are mutually dependent: the outline determines what should be searched next, newly retrieved evidence may reshape the outline, and report generation can expose gaps in support or structure that call for further refinement.

\section{Method}
\label{sec:method}

\subsection{ScaffoldAgent Overview}
\label{sec:overview}

\M~instantiates OEDR as a utility-guided outline optimization process. 
Starting from a root outline node initialized from the input question, \M~iteratively grows and refines an outline tree until it becomes a stable scaffold for long-form report generation. 
At iteration $t$, we denote the current outline tree as
\begin{equation}
\mathcal{T}_t=(\mathcal{V}_t,\mathcal{P}_t,\Phi_t),
\end{equation}
where $\mathcal{V}_t$ is the set of outline nodes, $\mathcal{P}_t$ denotes parent--child relations, and $\Phi_t$ stores node-level attributes. 
Each node $v\in\mathcal{V}_t$ corresponds to a section or subsection and is associated with
\begin{equation}
\Phi_t(v)=\big(h_v,\mathcal{E}_v,\theta_v\big),
\end{equation}
where $h_v$ is the node's intent, $\mathcal{E}_v$ is the supporting evidence and report segment attached to the node, and $\theta_v$ records accumulated utility statistics. 
The initial tree contains only the root node derived from the input question
$\mathcal{T}_0=\mathrm{Init}(q).
$

\M~comprises three specialized agents. 
The \textbf{Outline Agent} serves as the central controller: it maintains the evolving outline tree, selects which node to visit, and decides how the selected node should be updated. 
The \textbf{Search Agent} retrieves external evidence when an update requires new information. 
The \textbf{Reporter Agent} performs trial writing during optimization and produces the final report once the outline converges.

The optimization process is node-centric. 
At each iteration, the Outline Agent first selects a target node $v_t$ according to utility feedback, and then chooses one structural operation $o_t$ from:
\begin{equation}
o_t\in\{\text{Expansion},\text{Contraction},\text{Revision}\}.
\end{equation}
The selected operation updates the outline tree and yields utility feedback:
\begin{equation}
\mathcal{T}_{t+1}, U_t
=
\mathrm{Update}\big(\mathcal{T}_t,v_t,o_t;\pi_{\text{res}},\pi_{\text{rep}}\big),
\end{equation}
where $\pi_{\text{res}}$ and $\pi_{\text{rep}}$ denote the Search Agent and the Reporter Agent, respectively. 
Expansion and Revision invoke both the Search Agent and the Reporter Agent, as they require additional evidence and downstream writing feedback. 
Contraction reorganizes existing evidence and therefore only invokes the Reporter Agent to evaluate whether the compressed local structure remains coherent and grounded. 
The resulting utility feedback is used to update node statistics and guide subsequent iterations. 
When further refinement yields diminishing utility gains, \M~terminates the optimization and generates the final report from the converged scaffold.

\begin{figure*}[t]
  \centering
  \includegraphics[width=0.91\textwidth]{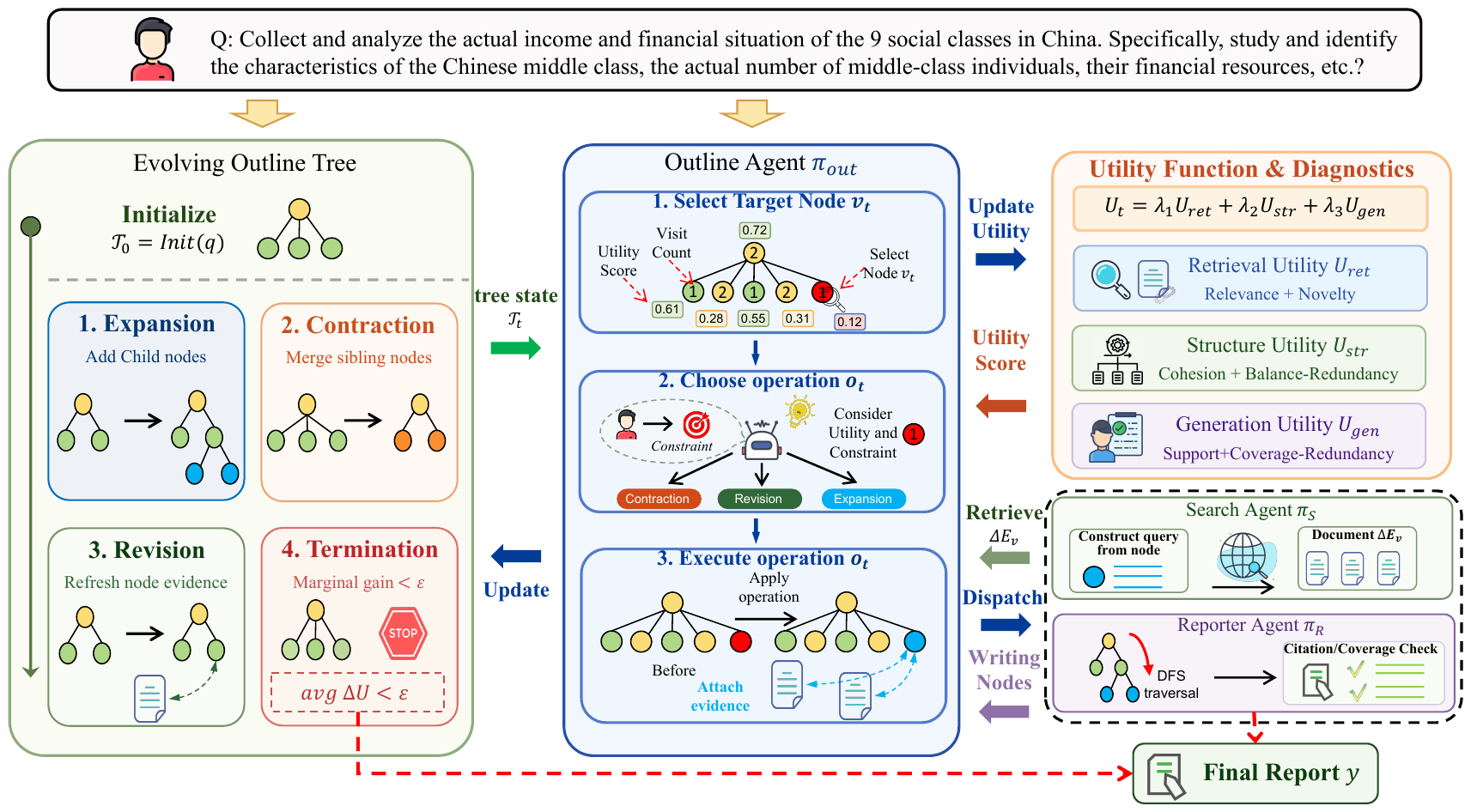}
  \captionsetup{font=footnotesize}
  \caption{
  Overview of \M. Starting from a root outline tree, the Outline Agent iteratively selects target nodes according to utility feedback and updates them through \emph{Expansion}, \emph{Contraction}, or \emph{Revision}. The Search Agent provides evidence when needed, and the Reporter Agent performs trial writing or final report generation. Utility feedback updates node statistics and guides optimization until the outline tree converges.
  }
  \label{fig:framework}
\end{figure*}
\subsection{Evolving Outline Tree}
\label{sec:tree}

The outline tree serves as the central state representation of \M. 
It starts from a root node derived from the input question and gradually evolves into a hierarchical report scaffold. 
Each node represents a potential section or subsection, carrying an intent description that specifies what the section should cover, a set of supporting evidence, and utility statistics accumulated from previous visits. 
By attaching evidence directly to outline nodes, \M~uses the outline not only as a writing plan but also as an evidence index that bridges retrieval and generation.

At each iteration, the Outline Agent must decide which node to visit next. 
An effective selection strategy should revisit low-quality nodes that require correction while still exploring under-visited parts of the tree to avoid early local bias. 
For each node $v$, it maintains utility statistics
\begin{equation}
\theta_v=(n_v,\bar{u}(v)),
\end{equation}
where $n_v$ is the number of visits and $\bar{u}(v)$ is the running mean utility of actions previously applied to this node. 
Let $\mathcal{V}_t$ denote the set of outline nodes at iteration $t$. 
The next target node is selected via a UCB-style rule:
\begin{equation}
\label{eq:ucb}
v_t=\arg\max_{v\in\mathcal{V}_t}
\left(
-\bar{u}(v)+c\sqrt{\frac{\ln N_t}{n_v}}
\right),
\end{equation}
where $N_t=\sum_{v\in\mathcal{V}_t}n_v$ is the total visit count and $c$ is the exploration coefficient. 
The first term prioritizes low-utility nodes that need improvement, while the second term encourages exploration of rarely visited nodes.

After an operation is applied to selected node, its statistics are updated with new utility $U_t$:
\begin{equation}
n_{v_t}\leftarrow n_{v_t}+1,
\bar{u}(v_t)\leftarrow \bar{u}(v_t)+
\frac{U_t-\bar{u}(v_t)}{n_{v_t}}.
\end{equation}
This incremental update allows the system to track which parts of the outline remain weak and which have already stabilized.

After a target node is selected, the Outline Agent determines how to update it based on its utility profile. 
A node with insufficient topical granularity typically triggers \emph{Expansion}; redundant or overlapping sibling nodes tend to trigger \emph{Contraction}; and a weakly supported or outdated node tends to prompt \emph{Revision}. 
Outline tree continues evolving until marginal utility gain falls below a threshold:
\begin{equation}
\frac{1}{k}\sum_{i=t-k+1}^{t}\Delta U_i<\epsilon,
\Delta U_i=U_i-U_{i-1}.
\end{equation}
This criterion indicates that further structural changes are unlikely to improve the final report. 
At this point, the Reporter Agent generates the final report from the converged outline.
 
\subsection{Outline Agent's Actions}
\label{sec:actions}

To mitigate scaffold drift (\textbf{C1}), \M~updates the outline through three explicit operations: \emph{Expansion}, \emph{Contraction}, and \emph{Revision}. 
These operations are inspired by belief revision theory~\cite{alchourron1985logic}, but serve here as practical structural actions rather than formal logical operators. 
Each addresses a common outline deficiency: insufficient granularity, redundant branching, or weak evidential grounding.

\begin{itemize}[leftmargin=*]
    \item \textbf{Expansion.}
    Expansion is applied when the selected node is too broad to support focused retrieval or detailed writing. 
    The Outline Agent decomposes the node into several finer-grained child nodes, each with a more specific intent. 
    Since newly created subtopics require supporting evidence, Expansion first invokes the Search Agent to retrieve relevant documents for each child node. 
    The Reporter Agent then performs trial writing on the expanded subtree to assess whether the new structure improves downstream generation quality.

    \item \textbf{Contraction.}
    Contraction is applied when sibling nodes are semantically overlapping, overly fragmented, or repeatedly supported by similar evidence. 
    The Outline Agent merges redundant branches into representative nodes and regenerates their intents to cover the consolidated content. 
    Since Contraction reorganizes existing evidence rather than introducing new subtopics, it does not invoke the Search Agent. 
    The Reporter Agent then verifies whether the compressed local structure remains coherent, non-redundant, and sufficiently grounded.

    \item \textbf{Revision.}
    Revision is applied when the selected node is weakly supported, outdated, or misaligned with its accumulated evidence. 
    The Outline Agent preserves the node's position in the tree but refreshes its intent and evidence grounding. 
    Revision first invokes the Search Agent to retrieve additional evidence and then the Reporter Agent to assess whether the revised node better supports grounded writing. 
    Unlike Expansion, Revision does not alter the size of the tree.
\end{itemize}

These three operations govern how the outline tree evolves throughout the research process. 
Expansion increases topical granularity, Contraction controls redundancy, and Revision strengthens local evidence grounding. 
Together, they allow \M~to maintain a scaffold that is stable enough for coherent writing while remaining adaptive to newly acquired evidence.

\subsection{Utility-Guided Feedback}
\label{sec:reward}

To address delayed feedback (\textbf{C2}), \M~uses utility as an inference-time control signal rather than a gradient-based training objective. 
After each node-level operation, the system evaluates the updated local scaffold from three perspectives: whether the operation acquires useful evidence, whether it improves structural organization, and whether it supports grounded writing. 
We denote the overall utility at iteration $t$ as:
\begin{equation}
U_t=\lambda_1U_{\text{ret}}+\lambda_2U_{\text{str}}+\lambda_3U_{\text{gen}},
\end{equation}
where $U_{\text{ret}}$, $U_{\text{str}}$, and $U_{\text{gen}}$ are retrieval, structure, and generation utility, respectively. 
The coefficients $\lambda_1$, $\lambda_2$, and $\lambda_3$ control relative importance.

\begin{itemize}[leftmargin=*]
    \item \textbf{Retrieval Utility.}
    $U_{\text{ret}}$ evaluates whether newly acquired evidence is both relevant to the target node and non-redundant with respect to existing evidence:
    \begin{equation}
    U_{\text{ret}}=
    w_{\text{rel}}\mathrm{Rel}
    +
    w_{\text{nov}}\mathrm{Nov}.
    \end{equation}
    Here, $\mathrm{Rel}$ measures semantic relevance between retrieved documents and the target node's intent, while $\mathrm{Nov}$ measures novelty relative to evidence already attached to the outline tree.

    \item \textbf{Structure Utility.}
    $U_{\text{str}}$ evaluates whether the outline scaffold remains coherent, balanced, and non-redundant:
    \begin{equation}
    U_{\text{str}}=
    w_{\text{coh}}\mathrm{Coh}
    +
    w_{\text{bal}}\mathrm{Bal}
    -
    w_{\text{red}}\mathrm{Red}.
    \end{equation}
    Here, $\mathrm{Coh}$ quantifies semantic consistency among related nodes, $\mathrm{Bal}$ measures structural balance, and $\mathrm{Red}$ penalizes redundant sibling branches.

    \item \textbf{Generation Utility.}
    $U_{\text{gen}}$ evaluates whether the current scaffold can support grounded writing. 
    The Reporter Agent performs trial writing on the affected node or subtree, and the resulting text is scored by:
    \begin{equation}
    U_{\text{gen}}=
    \alpha\rho_{\text{sup}}
    +
    \beta\rho_{\text{cov}}
    -
    \gamma\rho_{\text{red}},
    \end{equation}
    where $\rho_{\text{sup}}$ measures citation support, $\rho_{\text{cov}}$ measures alignment between node intent and generated content, and $\rho_{\text{red}}$ penalizes cross-section redundancy.
\end{itemize}

\subsection{Multi-Turn Scaffold Refinement}
  \label{sec:multiturn-refinement}

To support follow-up instructions without regenerating the entire report, \M extends the dynamic scaffold optimization process in two stages. \textbf{\ding{182} Instruction-Aware Outline Update.} Given a follow-up instruction, the Outline Agent receives it along with the scaffold retained from the previous turn. It first identifies the outline nodes affected by the new requirement and then updates the corresponding local subtrees. \emph{Expansion} introduces new subtopics required by the instruction, \emph{Revision} adjusts existing section intents, and \emph{Contraction} removes branches that become redundant after the update.
\textbf{\ding{183} Localized Agent Collaboration.}
The Outline Agent invokes the Search Agent when the modified nodes require additional evidence and engages the Reporter Agent to trial-write and evaluate the affected sections. Utility-guided optimization is confined to these local subtrees, while the structure, evidence, and content of unaffected sections are preserved. Once the local scaffold converges, the Reporter Agent regenerates the modified sections and integrates them with the retained content.

\subsection{Lightweight Implementation} \label{sec:lightweight-implementation} 
Repeated model-based utility estimation can increase inference overhead and introduce evaluator dependence. To reduce this cost, \textbf{ScaffoldAgent Lite} adopts two simplification strategies. 
\textbf{\ding{182} Hierarchical Candidate Pruning.} Before utility evaluation, the system reduces the candidate node--operation space at two levels. At the node level, it ranks outline nodes by their stored utility history and a UCB-style exploration bonus, retaining only the top-ranked target for evaluation in each iteration. At the operation level, inexpensive feasibility rules filter out currently inapplicable operations: \emph{Expansion} is restricted to leaf nodes, \emph{Revision} is prioritized for generic or insufficiently supported nodes, and \emph{Contraction} is considered only when sibling redundancy is detected. Pruning applies only to the current iteration. After each scaffold update, the candidate set is reconstructed, and a previously skipped candidate may be reconsidered if it becomes applicable under the updated structure or evidence. 
\textbf{\ding{183} Lightweight Utility Estimation.} For the remaining candidates, the three utility dimensions are estimated from local signals. Retrieval utility considers evidence coverage and the amount of non-redundant information introduced relative to cached evidence. Structure utility considers tree depth, branching, and overlap among sibling titles. Generation utility relies on a limited local trial draft to estimate evidence support, intent coverage, and repetition. All three dimensions can be computed efficiently from local statistics and lightweight similarity measures.


\section{Experiments}
We evaluate \M~against the following research questions:
\begin{itemize}[leftmargin=*,noitemsep,topsep=2pt]
    \item \textbf{RQ1:} Does \M~outperform existing methods in open-ended deep research?
    \item \textbf{RQ2:} How much does each internal component of \M{}---the three outline actions (\emph{Expansion}, \emph{Contraction}, \emph{Revision}) and the three utility dimensions ($U_{\text{ret}}$, $U_{\text{str}}$, $U_{\text{gen}}$) that score them contribute to the final performance?
    \item \textbf{RQ3:} Can \M~handle multi-turn follow-ups by incrementally updating a previously built outline tree?
\end{itemize}
\subsection{Experimental Setup}
\textbf{\ding{182} Evaluation Benchmarks.}
 We evaluate \M~on two open-ended deep-research benchmarks: \textbf{DeepResearch Bench}~\cite{du2025deepresearch} and \textbf{DeepResearch Gym}~\cite{coelho2025deepresearchgym}. The former assesses report quality via RACE (Overall, Comprehensiveness, Insight, Instruction-following, Readability) and factual grounding via FACT (effective citation rate Eff.c.\ and citation accuracy C.acc.). The latter scores six report dimensions: Clarity, Depth, Balance, Breadth, Supportability, and Insightfulness. Further details are deferred to Appendix~\ref{appendix:datasets}. 
 
 \textbf{\ding{183} Baselines.} 
 We compare \M~against baselines from three paradigms: \emph{Naive} (direct Prompt and RAG), \emph{Single-Agent} (ReAct~\cite{yao2023react}, IRCoT~\cite{trivedi2023interleaving}, WebShaper~\cite{tao2025webshaper}), and \emph{Multi-Agent} (STORM~\cite{shao2024assisting}, WebWeaver~\cite{li2025webweaver}, EDR~\cite{prabhakar2025enterprise}, StackPlanner~\cite{zhang2026stackplanner}). Further details are deferred to Appendix~\ref{appendix:baselines}.  
 
 \textbf{\ding{184} Implementation Details.} 
 We use \textbf{DeepSeek-V3.2}~\cite{liu2024deepseek} and \textbf{Qwen3-32B}~\cite{yang2025qwen3} as backbone LLMs, and adopt Bocha as the web-search API. Further implementation details are deferred to Appendix~\ref{appendix:implementation}.

\subsection{Main Result Analysis (RQ1)}
\label{sec:main-results}

\begin{table*}[t]
\centering
\fontsize{7.5pt}{8pt}\selectfont
\setlength{\tabcolsep}{2.4pt}
\renewcommand{\arraystretch}{1.25}

\resizebox{\textwidth}{!}{
\begin{tabular}{l l | c c c c c | c c | c c c c c | c c}
\toprule
\rowcolor{gray!30}
\multicolumn{2}{c|}{\textbf{}} &
\multicolumn{7}{c|}{\textbf{Qwen3-32B}} &
\multicolumn{7}{c}{\textbf{DeepSeek-V3.2}} \\

\rowcolor{gray!30}
& &
\multicolumn{5}{c}{\textbf{RACE}} & \multicolumn{2}{c|}{\textbf{FACT}} &
\multicolumn{5}{c}{\textbf{RACE}} & \multicolumn{2}{c}{\textbf{FACT}} \\

\rowcolor{gray!30}
\textbf{Paradigm} & \textbf{Approach} &
\textbf{Over.} & \textbf{Comp} & \textbf{Ins.} & \textbf{Inst} & \textbf{Read} &
\textbf{Eff.c.} & \textbf{C.acc.} &
\textbf{Over.} & \textbf{Comp} & \textbf{Ins.} & \textbf{Inst} & \textbf{Read} &
\textbf{Eff.c.} & \textbf{C.acc.} \\
\midrule

\multirow{2}{*}{Naive}
& Prompt
  & 39.54 & 36.87 & 35.25 & 45.34 & 43.67 & --    & --
  & 43.68 & 41.77 & 41.67 & 47.21 & 46.35 & --  & --  \\  
  
& RAG
  & 36.68 & 34.49 & 33.35 & 43.07 & 42.92 & 2.93  & 10.86
  & 39.50 & 36.46 & 37.19 & 43.35 & 42.83 & 5.51  & 16.18  \\
\midrule

\multirow{3}{*}{Single-Agent}
& ReAct
  & 42.21 & 40.66 & 38.65 & 46.10 & 44.56 & 4.69  & 23.78
  & 43.26 & 42.70 & 40.53 & 46.10 & 46.43 & 4.15  & 17.83  \\
& IRCoT
  & 42.46 & 40.70 & 39.55 & 46.23 & \textbf{45.70} & 15.77 & 40.04
  & 39.73 & 36.80 & 36.81 & 44.57 & 40.70 & 34.83 & 75.18 \\
& WebShaper
  & 34.93 & 31.58 & 26.17 & 44.81 & 40.38 & --    & --
  & 34.93 & 31.58 & 26.17 & 44.81 & 40.38 & -- & --\\
\midrule

\multirow{5}{*}{Multi-Agent}
& STORM
  & 32.81 & 30.63 & 31.66 & 34.88 & 36.90 & 15.18 & 36.90
  & 35.26    & 33.69    & 31.52   & 39.02   &39.07   & 17.42    & 38.59 \\
& WebWeaver
  & 41.28 & 39.88 & 39.68 & 44.65 & 39.69 & 8.42 & 32.72
  & 43.52     & 41.27   & 42.37    &47.18    &44.91    &39.00    & \textbf{88.35} \\
& EDR
  & 40.99 & 39.49 & 34.61 & 45.22 & 39.65 & 18.78 & 23.70
  & 43.13    & 41.95    & 40.27    & 46.24   &44.50  & 24.28    &34.79 \\
& StackPlanner
  & 41.55 & 39.84 & 37.79 & 46.41 & 44.79  & 9.63  & 26.50
  & 46.82    & 46.42    & 46.29    & 48.25    & 46.75    & 32.75    & 50.40 \\
\cmidrule{2-16}
\rowcolor[HTML]{F0F6FF}
& \textbf{\M}
  & \textbf{44.70} & \textbf{43.54} & \textbf{45.25} & \textbf{46.49} & 42.85
  & \textbf{30.42} & \textbf{54.32}
  & \textbf{48.27}    & \textbf{46.40}    & \textbf{49.67 }    & \textbf{48.98}    & \textbf{47.31}
  & \textbf{51.18}    & 62.20 \\

\bottomrule
\end{tabular}
}
\caption{Report-generation results on \textbf{DeepResearch Bench} using
Qwen3-32B and DeepSeek-V3.2. RACE measures report quality and FACT evaluates
factual grounding (effective citation rate, Eff.c.; citation accuracy,
C.acc.). Best results per backbone are in \textbf{bold}. ``--'' indicates
unavailable results.}
\label{tab:main_drb}
\end{table*}

\begin{table*}[t]
\centering
\fontsize{8pt}{9pt}\selectfont
\setlength{\tabcolsep}{10pt}
\renewcommand{\arraystretch}{1.25}

\begin{tabular}{l l | c c c c c c | c}
\toprule
\rowcolor{gray!30}
\textbf{Paradigm} & \textbf{Approach} &
\textbf{Cla} & \textbf{Depth} & \textbf{Bal} & \textbf{Brea} & \textbf{Sup} & \textbf{Ins} & \textbf{Avg.} \\
\midrule

\multirow{2}{*}{Naive}
& Prompt & \textbf{70.40} & 69.10 & \textbf{81.10} & 76.70 & 0.00 & 53.20 & 58.42 \\
& RAG    & 64.60 & 67.20 & 74.60 & 71.00 & 4.30 & 52.40 & 55.68 \\
\midrule
\multirow{3}{*}{Single-Agent}
& ReAct     & 60.50 & \textbf{83.40} & 60.30 & 74.90 & 61.90 & 65.40 & 67.73 \\
& IRCoT     & 62.30 & 73.10 & 74.60 & 77.30 & 83.80 & 56.60 & 71.28 \\
& WebShaper & 49.40 & 59.00 & 54.20 & 75.20 & 26.80 & 59.70 & 54.05 \\
\midrule
\multirow{5}{*}{Multi-Agent}
& STORM        & 31.00 & 54.00 & 67.00 & 65.50 & 81.50 & 46.00 & 57.50 \\
& WebWeaver    & 62.10 & 77.90 & 71.20 & 80.50 & 59.50 & 67.10 & 69.72 \\
& EDR          & 55.80 & 76.80 & 74.60 & 84.90 & \textbf{89.50} & 62.70 & 74.05 \\
& StackPlanner & 62.10 & 69.40 & 66.10 & 77.80 & 31.80 & 59.20 & 61.07 \\
\cmidrule{2-9}
\rowcolor[HTML]{F0F6FF}
& \textbf{\M}
  & 63.30 & 82.40 & 75.50 & \textbf{85.20} & 79.80 & \textbf{68.80}
  & \textbf{75.83} \\

\bottomrule
\end{tabular}
\caption{Multi-dimensional evaluation on \textbf{DeepResearch Gym} under the
Qwen3-32B backbone. Dimensions: Clarity, Depth, Balance, Breadth, Support,
Insight. Best in \textbf{bold}.}
\label{tab:main_gym}
\end{table*}

We evaluate \M~on \textit{DeepResearch Bench}
(Table~\ref{tab:main_drb}, both Qwen3-32B~\cite{yang2025qwen3} and DeepSeek-V3.2 backbones)
and \textit{DeepResearch Gym} (Table~\ref{tab:main_gym}, Qwen3-32B).

\paragraph{Comparison with Baselines.}
\M~achieves the strongest Qwen3-32B results on DeepResearch Bench,
reaching 44.70 RACE Overall and improving over the best baseline,
IRCoT, by $+2.24$ points. The gain is accompanied by stronger factual
grounding: \M~obtains the highest Eff.c.\ (30.42) and C.acc.\ (54.32),
whereas the strongest single-agent baselines remain substantially lower in citation accuracy (ReAct 23.78 and IRCoT 40.04). Multi-agent baselines do not close this gap: WebWeaver is competitive in RACE. Overall (41.28) but trails by 21.60 points in C.acc., while EDR and STORM score below 41 RACE Overall and below 37 C.acc. These results show that \M~improves report quality without sacrificing grounding.

\paragraph{Robustness across Backbones.}
On DeepSeek-V3.2, \M~remains the best method in RACE Overall, reaching
48.27 and surpassing StackPlanner by $+1.45$ points. It also ranks first
on all five RACE sub-dimensions, indicating that the improvement is not
driven by a single aspect of report quality. Baseline behavior changes
substantially across backbones: IRCoT drops from 42.46 to 39.73 RACE
Overall, while StackPlanner rises from 41.55 to 46.82. In contrast, \M
improves monotonically on every Bench metric (RACE 44.70$\to$48.27,
Eff.c.\ 30.42$\to$51.18, C.acc.\ 54.32$\to$62.20), suggesting that the
utility-guided framework scales with stronger backbone capability rather
than overfitting to a particular model.

\paragraph{Multi-Dimensional Evaluation on Gym.}
On DeepResearch Gym, \M~achieves the highest average score, 75.83, outperforming the strongest baseline EDR by $+1.78$ points. It leads two of the six dimensions, Breadth (85.20), and Insight (68.80), which is consistent with its explicit outline expansion and refinement process. Although EDR obtains the highest Supportability score (89.50), \M~remains competitive on Supportability (79.80) and keeps the remaining dimensions close to the best baseline, showing that its gains in depth and insight do not come from degrading overall structural quality.

\subsection{Component Analysis}
\begin{figure}[t]
\centering
\includegraphics[width=\linewidth]{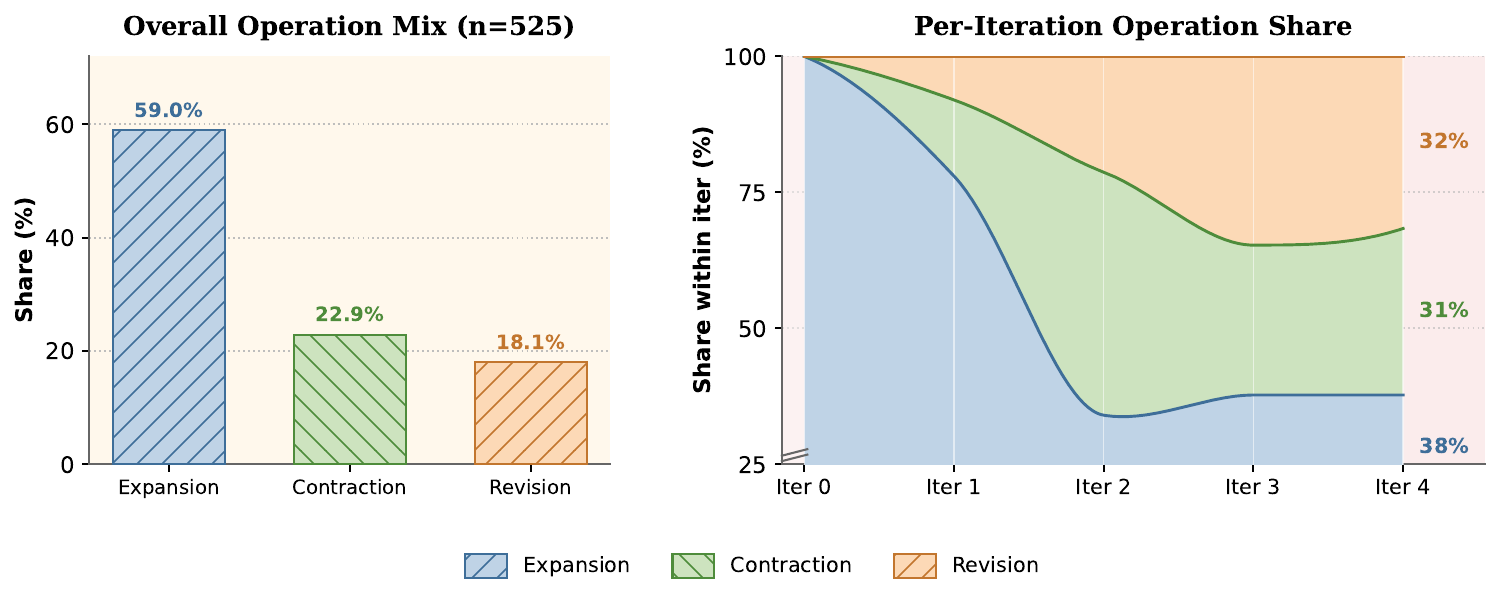}
\caption{Action behavior analysis. Left: overall distribution of planning actions. Right: action frequency across outline
refinement iterations.}
\label{fig:action_dist}
\end{figure}

\textbf{\ding{182} Outline-Action Behavior and Ablation.}
For \textbf{RQ2} (action side), Table~\ref{tab:ablation-actions} ablates the three AGM-aligned outline actions.
Removing \textit{Contraction} lets the tree widen unchecked, dropping RACE 48.27$\to$44.72 ($-3.55$) and C.acc.\ 62.20$\to$57.79. Removing \textit{Revision} blocks evidence refresh and gives the lowest single-action RACE (43.09, $-5.18$); C.acc.\ slips only to 58.11 because the retrieved evidence still locally supports the (ill-posed) claims it was fetched for citations stay attributable, but the underlying arguments are weaker, which RACE penalizes and FACT does not. 
Disabling both (\textit{Expansion-only}) yields suboptimal results on all three metrics (RACE 41.73, Eff.c.\ 45.20, C.acc.\ 59.32),
and we further observe non-terminating runs in which the outline keeps inflating until the budget is exhausted, confirming that the two shrinking actions are load-bearing. Figure~\ref{fig:action_dist} corroborates this from the behavior side: \textit{Expansion} dominates the early budget, then \textit{Contraction} and \textit{Revision} take over for merging and refresh, with all three active throughout.

\begin{table}[t]
\centering
\fontsize{9pt}{9pt}\selectfont
\setlength{\tabcolsep}{4pt}
\renewcommand{\arraystretch}{1.1}

\begin{tabular}{l|c|cc}
\toprule
\rowcolor{gray!10}
\textbf{Variant} & \textbf{RACE} & \textbf{Eff.c.} & \textbf{C.acc.} \\
\midrule

\textbf{\M~(Full)} & \textbf{48.27} & \textbf{51.18} & \textbf{62.20} \\

\midrule

w/o Contraction       & 44.72 & 43.76 & 57.79 \\
w/o Revision          & 43.09 & 50.00 & 58.11 \\
Expansion-only        & 41.73 & 45.20 & 59.32 \\

\bottomrule
\end{tabular}
\caption{Ablation on the outline-action space.}
\label{tab:ablation-actions}
\end{table}

\begin{figure*}[t]
\centering
\includegraphics[width=0.85\textwidth]{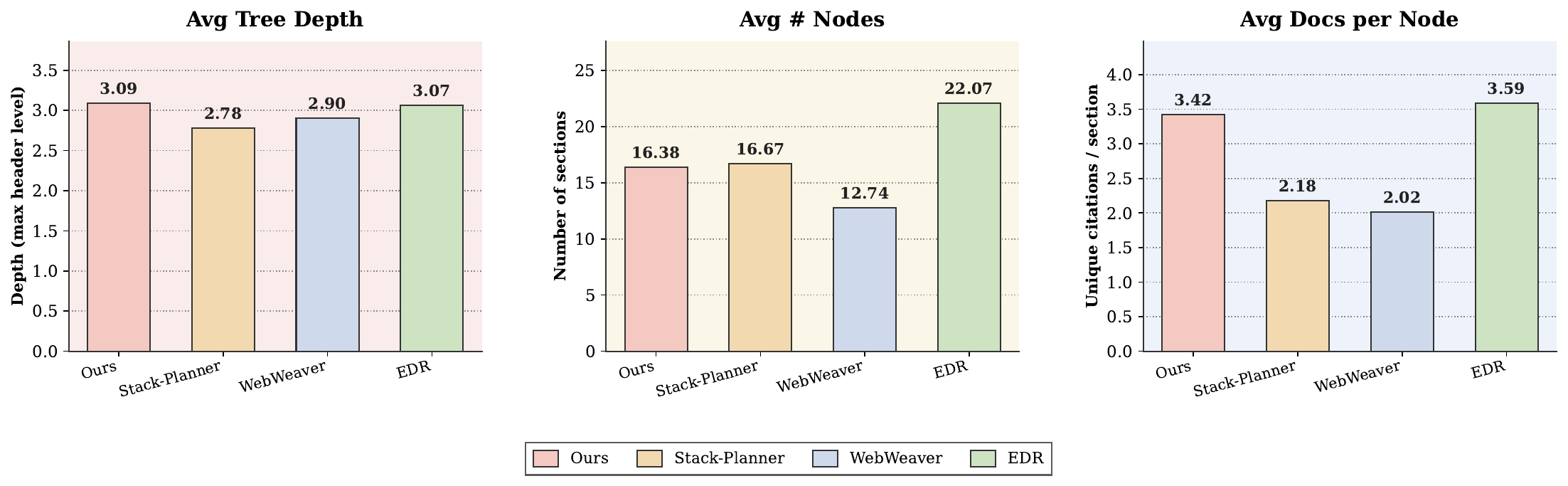}
\caption{Outline tree statistics across methods. Left: average tree depth. Middle: average number of nodes. Right: average documents per node.(DeepResearch Gym)}
\label{fig:tree_stats}
\end{figure*}

\textbf{\ding{183} Utility-Dimension Ablation.}
For \textbf{RQ2} (utility side), Table~\ref{tab:ablation-utility} isolates the three utility terms used to score candidate actions. Removing $U_{\text{ret}}$ mainly weakens evidence supply, sharply reducing Eff.c.\ 51.18$\to$39.42 ($-11.76$) while C.acc.\ drops more mildly 62.20$\to$55.68, suggesting that retrieval utility primarily controls whether the tree acquires enough relevant, non-redundant evidence. Removing $U_{\text{str}}$ disrupts global outline organization, dropping RACE 48.27$\to$44.17 ($-4.10$) with only moderate FACT degradation (Eff.c.\ 47.86, C.acc.\ 58.41); local claims may remain attributable, but redundant or unbalanced branches weaken comprehensiveness, insight, and readability. Removing $U_{\text{gen}}$ is most damaging overall, yielding the lowest RACE (43.58, $-4.69$) and the largest C.acc.\ drop 62.20$\to$50.82 ($-11.38$), since trial writing directly scores citation support, intent coverage, and redundancy. Together, these results show that $U_{\text{ret}}$ supplies evidence, $U_{\text{str}}$ regularizes the scaffold, and $U_{\text{gen}}$ aligns outline optimization with downstream report generation.

\begin{table}[t]
\centering
\fontsize{9pt}{9pt}\selectfont
\setlength{\tabcolsep}{4pt}
\renewcommand{\arraystretch}{1.1}

\begin{tabular}{l|c|cc}
\toprule
\rowcolor{gray!10}
\textbf{Variant} & \textbf{RACE} & \textbf{Eff.c.} & \textbf{C.acc.} \\
\midrule

\textbf{\M~(Full)} & \textbf{48.27} & \textbf{51.18} & \textbf{62.20} \\

\midrule

w/o $U_{\text{ret}}$  & 45.42 & 39.42 & 55.68 \\
w/o $U_{\text{str}}$  & 44.17 & 47.86 & 58.41 \\
w/o $U_{\text{gen}}$  & 43.58 & 44.73 & 50.82 \\

\bottomrule
\end{tabular}
\caption{Ablation on the utility dimensions.}
\label{tab:ablation-utility}
\end{table}

\textbf{\ding{184} Multi-turn Interaction.}
For \textbf{RQ3}, Table~\ref{tab:multiturn} compares three multi-turn refinement strategies. Dataset construction and evaluation details are provided in Appendix~\ref{app:multiturn-data}. \emph{ReAct-FW} starts a new research process from the combined original and follow-up queries, whereas \emph{ReAct-GR} globally rewrites the first-turn report according to the follow-up instruction. \M instead updates the retained scaffold through localized \emph{Expansion}, \emph{Revision}, and \emph{Contraction} while preserving unaffected content. \M obtains a total score of 72.60, outperforming ReAct-FW (56.75) and ReAct-GR (59.70) by 15.85 and 12.90 points, respectively. These results indicate that localized scaffold refinement more effectively incorporates follow-up instructions while retaining prior content than either fresh research or global rewriting.

 \begin{table}[t]
 \centering
 \fontsize{7.5pt}{8.5pt}\selectfont
 \setlength{\tabcolsep}{2.4pt}
 \renewcommand{\arraystretch}{1.1}
 
 \begin{tabular}{l|ccccc|c}
 \toprule
 \rowcolor{gray!10}
 \textbf{Setting} & \textbf{Ins.}$\uparrow$ & \textbf{Evi.}$\uparrow$ & \textbf{Log.}$\uparrow$ & \textbf{Com.}$\uparrow$ & \textbf{Exp.}$\uparrow$ & \textbf{Tot.}$\uparrow$ \\
 \midrule
 ReAct-FW      & 13.80 & 12.20 & 12.75 &  6.60 & 11.40 & 56.75 \\
 ReAct-GR      & 12.65 & 10.85 & 11.55 & 10.90 & 13.75 & 59.70 \\
 \midrule
 \textbf{\M}   & \textbf{15.35} & \textbf{13.40} & \textbf{14.55} & \textbf{15.10} & \textbf{14.20} & \textbf{72.60} \\
 \bottomrule
 \end{tabular}
 \caption{Multi-turn refinement on 100 follow-up instances from DeepResearch Bench. Each dimension is scored 0--20; Tot.\ is their sum.}
 \label{tab:multiturn}
 \end{table}

\paragraph{\ding{185} Outline Tree Statistics.}


Figure~\ref{fig:tree_stats} compares the outline structures produced by different methods. \M~yields the deepest trees (3.09) while keeping the node count moderate (16.38), close to StackPlanner (16.67) and well below EDR (22.07), suggesting that Contraction controls tree growth. More importantly, \M~achieves the highest evidence density, with 5.85 documents per node versus 3.59 for EDR, 2.18 for StackPlanner, and 2.02 for WebWeaver. Thus, \M~does not simply expand the outline; it builds a compact scaffold with richer evidence support for each section.




\section{Conclusion}

\M~presents a utility-guided dynamic outline optimization framework for open-ended deep research. Instead of treating the outline as a static pre-writing plan, \M~maintains it as an evolving scaffold that coordinates evidence retrieval, content organization, and long-form report generation. By modeling outline evolution through three explicit operations, Expansion, Contraction, and Revision, the framework enables controllable structural updates that incorporate new information, reduce redundant branches, and refresh weakly supported sections. \M~further introduces a utility-guided feedback mechanism that jointly considers retrieval gain, structural quality, and trial-generation quality, providing inference-time signals for node selection, operation scheduling, and termination. 
Experiments on DeepResearch Bench and DeepResearch Gym show that \M~achieves strong overall performance in report quality and factual grounding compared with existing deep research agents, demonstrating the effectiveness of dynamic outline optimization for open-ended research tasks.

\section*{Acknowledgments}
This work is supported by the Natural Science Foundation of China (Grant No.62506010) 

\section*{Limitations}

Despite the consistent improvements achieved by \M, several directions remain for future work. First, our multi-turn evaluation is still preliminary. Current open-ended deep research benchmarks are largely single-turn, whereas realistic use often involves longer dialogues in which users iteratively revise the scope, constraints, or subtopics. Future work could develop longer multi-turn trajectories and more comprehensive evaluation protocols for such interactive research scenarios.
Second, due to the substantial cost of large-scale deep research evaluation, our experiments are conducted primarily with widely used open-source foundation models and a fixed search interface, rather than the strongest available models and search systems. Future work may further investigate how the framework scales when combined with more capable backbone models, advanced search infrastructures, and larger retrieval budgets.
Finally, \M~currently relies on utility-based inference-time control without explicitly learning an outline-evolution policy. While this design is lightweight and effective, dynamic outline optimization naturally forms a sequential decision-making problem. A promising direction for future work is to learn such policies with reinforcement learning, potentially improving outline expansion, refinement, and information gathering throughout the report generation process.

\section*{Ethical considerations}
All experiments in this work were carried out on publicly available open-ended deep research benchmarks, including \textit{DeepResearch Bench} and \textit{DeepResearch Gym}, following their respective licenses and usage terms. The study does not involve human or animal subjects, and we did not collect, use, or disclose any personally identifiable information.
\bibliography{custom}

\appendix

\newpage

\section{More Method}
\label{sec:algorithm}
\paragraph{Optimization Algorithm.}
 Algorithm~\ref{alg:more-method} summarizes the inference-time outline optimization (Opt.) process of \M.

  \begin{algorithm}[t]
\caption{Outline Opt. in \M}
 \label{alg:more-method}
 \begin{algorithmic}[1]
 \Require Query $q$, search interface $\mathcal{S}$, max steps $T$, threshold $\epsilon$
 \Ensure Final report $y$
 \State Initialize outline tree $\mathcal{T}_0$ from $q$
 \State Initialize node statistics $\theta_v=(n_v,\bar{u}_v)$
 \For{$t=0$ to $T-1$}
     \State Select $v_t$ using historical utility feedback
     \State Choose $o_t \in \{\textbf{Expansion},\textbf{Contraction},\textbf{Revision}\}$
     \If{$o_t \neq \textsc{Contraction}$}
         \State Retrieve evidence $\Delta\mathcal{E}_{v_t}$ via $\mathcal{S}$
     \EndIf
     \State Update outline: $\mathcal{T}_{t+1}\leftarrow o_t(\mathcal{T}_t,v_t)$
     \State Trial-write the affected subtree
     \State Score updated subtree by $U_{\text{ret}},U_{\text{str}},U_{\text{gen}}$
     \State Compute $U_t=\lambda_1U_{\text{ret}}+\lambda_2U_{\text{str}}+\lambda_3U_{\text{gen}}$
     \State Update $\theta_{v_t}$ with $U_t$
     \If{recent utility gain $<\epsilon$}
         \State \textbf{break}
     \EndIf
 \EndFor
 \State Generate $y$ along the final outline
 \State \Return $y$
 \end{algorithmic}
 \end{algorithm}

\paragraph{Utility Signals and Computation.}
We provide further details on how the utility components in Section~\ref{sec:reward} are computed. The overall utility is computed as
\begin{equation}
U_t=\lambda_1U_{\text{ret}}+\lambda_2U_{\text{str}}+\lambda_3U_{\text{gen}},
\end{equation}
where retrieval utility, structure utility, and generation utility capture complementary aspects of the updated local scaffold.

 \textbf{Retrieval utility.}
 After an action retrieves a set of new documents
 $\Delta \mathcal{E}_v=\{d_1,\ldots,d_k\}$ for the target node $v$,
 we compute a retrieval reward following the intuition of maximal marginal relevance (MMR)~\cite{carbonell1998use}:
 useful evidence should be relevant to the current information need while adding non-redundant information.
 Formally, we define
 \begin{equation}
 U_{\text{ret}} =
 w_{\text{rel}}\mathrm{Rel}+
 w_{\text{nov}}\mathrm{Nov},
 \end{equation}
 where $w_{\text{rel}}$ and $w_{\text{nov}}$ control the trade-off between relevance and novelty.
 
 $\mathrm{Rel}$ measures semantic alignment between the retrieved documents and the target node intent.
 Let $q_v$ denote the retrieval query derived from node $v$, including its title and local outline context.
 Given the embedding of $q_v$ and each retrieved document $d_j$, relevance is computed as
 \begin{equation}
 \mathrm{Rel}
 =
 \frac{1}{k}\sum_{j=1}^{k}
 \mathrm{sim}(q_v,d_j),
 \end{equation}
 where $\mathrm{sim}(\cdot,\cdot)$ denotes cosine similarity in the embedding space.
 
 $\mathrm{Nov}$ measures the marginal information gain of the retrieved documents with respect to cached evidence in the current outline.
 For each new document $d_j$, we compare it with the existing evidence set $\mathcal{E}_{\mathrm{cache}}$ attached to the target node and related outline nodes:
 \begin{equation}
 \mathrm{Nov}
 =
 \frac{1}{k}\sum_{j=1}^{k}
 \left(
 1-\max_{d'\in\mathcal{E}_{\mathrm{cache}}}
 \mathrm{sim}(d_j,d')
 \right).
 \end{equation}
 This term assigns higher scores to documents that introduce information not already covered by the current scaffold, thereby encouraging broader evidence acquisition rather than repeated retrieval of similar passages.

 \textbf{Structure utility.}
 Structure utility evaluates whether the updated outline remains coherent, balanced, and non-redundant:
 \begin{equation}
 U_{\text{str}}=w_{\text{coh}}\mathrm{Coh}+w_{\text{bal}}\mathrm{Bal}-w_{\text{red}}\mathrm{Red}.
 \end{equation}
 $\mathrm{Coh}$ is estimated by an LLM judge that assesses whether the child intents are semantically consistent with the parent intent and whether the subtree forms a coherent decomposition of the target topic.
 
 $\mathrm{Bal}$ measures whether the updated local subtree has an appropriate depth and branching structure.
 Let $\mathcal{T}_v$ denote the subtree rooted at node $v$, and let $b_i$ be the number of children of node $i$ in this subtree.
 We define
 \begin{equation}
 \mathrm{Bal}
 =
 \exp(-\Delta_d-\Delta_b),
 \end{equation}
 where
 $\Delta_d=|\mathrm{depth}(\mathcal{T}_v)-d^\star|/d^\star$ measures the deviation from the desired local depth $d^\star$, and
 $\Delta_b=\mathrm{Var}(\{b_i\}_{i\in\mathcal{T}_v})/(b^\star)^2$ measures the imbalance of the branching distribution with respect to the desired branching factor $b^\star$.
 This score is high when the subtree has a reasonable depth and a stable branching pattern, and decreases when the subtree becomes overly shallow, overly deep, or structurally imbalanced.
 
 $\mathrm{Red}$ measures redundancy among sibling nodes.
 We estimate it using pairwise semantic similarity between sibling intents and their attached evidence, with an LLM redundancy judgment for near-duplicate subtopics when similarity alone is ambiguous.

\textbf{Generation utility.}
Generation utility evaluates whether the updated local scaffold can support grounded writing:
\begin{equation}
U_{\text{gen}}=\alpha\rho_{\text{sup}}+\beta\rho_{\text{cov}}-\gamma\rho_{\text{red}}.
\end{equation}
The Reporter Agent first performs local trial writing on the affected node or subtree. We then score the generated draft from three aspects. $\rho_{\text{sup}}$ measures factual support. We split the trial draft into claim units and use an NLI model to estimate whether each claim is entailed by the evidence attached to the target node:
\begin{equation}
\rho_{\text{sup}}
=\frac{1}{|\mathcal{C}|}\sum_{c\in\mathcal{C}}
\max_{e\in\mathcal{E}_v}p_{\text{NLI}}(\mathrm{entail}\mid e,c),
\end{equation}
where $\mathcal{C}$ is the set of claims extracted from the trial draft. $\rho_{\text{cov}}$ measures intent coverage. We ask an LLM judge to compare the node intent with the generated local draft and return a normalized score indicating whether the draft covers the required aspects. $\rho_{\text{red}}$ penalizes repeated content in the trial draft and semantic overlap with neighboring sections, estimated by semantic similarity and an LLM redundancy check.

\begin{table*}[t]
\centering
\small
\begin{tabular}{l l l}
\toprule
\textbf{Component} & \textbf{Signal} & \textbf{Estimator} \\
\midrule
$\mathrm{Rel}$ & Evidence--intent relevance & Semantic similarity / LLM judge \\
$\mathrm{Nov}$ & Evidence novelty & Max similarity to cached evidence \\
$\mathrm{Coh}$ & Subtree coherence & LLM judge \\
$\mathrm{Bal}$ & Tree balance & Depth and branching statistics \\
$\mathrm{Red}$ & Sibling redundancy & Similarity / LLM judge \\
$\rho_{\text{sup}}$ & Claim support & NLI entailment model \\
$\rho_{\text{cov}}$ & Intent coverage & LLM judge \\
$\rho_{\text{red}}$ & Draft redundancy & Similarity / LLM judge \\
\bottomrule
\end{tabular}
\caption{Utility components and their estimators.}
\label{tab:utility-estimators}
\end{table*}

\section{Experiment Datasets}
\label{appendix:datasets}

We evaluate our approach on two widely used benchmarks targeting end-to-end deep research agent evaluation. Key statistics are summarized in Table~\ref{tab:dataset_overview}.
\paragraph{\ding{182} DeepResearch Bench~\cite{du2025deepresearch}.} 
A benchmark of 100 PhD level tasks (50 EN and 50 ZH) across 22 expert curated domains, with the topical distribution calibrated against 96{,}147 real world user queries. Evaluation combines \textbf{RACE}, an LLM judge protocol that scores reports against a reference along adaptively weighted dimensions (comprehensiveness, insight, instruction following, readability), and \textbf{FACT}, which verifies citation and claim pairs to report citation accuracy (C.\,Acc.) and effective citations per task (E.\,Cit.).

\paragraph{\ding{183} DeepResearch Gym~\cite{coelho2025deepresearchgym}.} 
An open-source sandbox built on top of the Researchy Questions dataset. Following the released pipeline, we run our system on a 100-query subset sampled by document-click frequency, where higher-clicked queries naturally elicit multi-source exploration and long-form synthesis. Each report is then scored under an LLM-as-a-judge protocol along six complementary dimensions: \textbf{Clarity}, \textbf{Depth}, \textbf{Balance}, \textbf{Breadth}, \textbf{Support} and \textbf{Insightfulness}.

\begin{table}[ht]
\centering
\small
\setlength{\tabcolsep}{6pt}
\renewcommand{\arraystretch}{1.15}
\begin{tabular}{lcc}
\toprule
\rowcolor{gray!10}
\textbf{Dataset} & \textbf{\#Samples} & \textbf{Lang.} \\
\midrule
DeepResearch Bench & 100 & EN / ZH \\
DeepResearch Gym    & 100 & EN \\
\bottomrule
\end{tabular}
\caption{Overview of deep research benchmarks.}
\label{tab:dataset_overview}
\end{table}

\paragraph{\ding{184} Multi-Turn Refinement Extension.}
\label{app:multiturn-data}
To evaluate follow-up interaction, we construct a multi-turn extension based on DeepResearch Bench. For each sampled first-turn query~$q$ and its corresponding report~$y$, we generate follow-up instructions~$q'$ that cover common ways users refine a deep research report: asking for deeper explanation of a local point, requesting additional evidence or citations, adding a new comparison target or constraint, adjusting the presentation, or correcting an erroneous definition or premise. This yields triples $(q,y,q')$ that test whether an agent can revise an existing report without unnecessarily changing unaffected content. We evaluate the updated reports with an LLM judge on five 0--20 dimensions: \textbf{Ins.} for follow-up instruction satisfaction, \textbf{Evi.} for evidence support of new or modified claims, \textbf{Log.} for consistency with the surviving report, \textbf{Com.} for preserving important first-turn content while completing the follow-up, and \textbf{Exp.} for clarity and rigor of expression. \textbf{Tot.} is the sum of the five scores.

\section{Baseline Implementation Details}
\label{appendix:baselines}

We compare our method against baselines drawn from three paradigms
(\emph{Naive}, \emph{Single-Agent}, and \emph{Multi-Agent}),
covering distinct strategies for reasoning, retrieval, and agent coordination.
Implementation details for each paradigm are described below.

\paragraph{\ding{182} Naive.}
Naive baselines do not involve explicit agentic reasoning or coordination mechanisms. They either rely solely on the LLM's parametric knowledge or incorporate retrieval in a fixed, heuristic manner.
\begin{itemize}
    \item \textbf{Prompt.}
    A non-retrieval baseline where the LLM directly generates the final report from the input query using only its parametric knowledge, without any external evidence.
    \item \textbf{RAG.}
    A retrieval-augmented baseline that issues the input query to the search API once, concatenates the top-ranked passages with the query, and prompts the LLM to produce the report in a single pass.
\end{itemize}

\paragraph{\ding{183} Single-Agent.}
Single-Agent baselines rely on a single LLM that alternates between reasoning and tool use through prompting, without explicit coordination among multiple agents.

\begin{itemize}
    \item \textbf{ReAct~\cite{yao2023react}.}
    ReAct interleaves reasoning and tool invocation in a single loop, enabling the model to iteratively retrieve information and update its reasoning before producing the final answer.

    \item \textbf{IRCoT~\cite{trivedi2023interleaving}.}
    IRCoT alternates retrieval and chain-of-thought reasoning, using intermediate reasoning steps to formulate new retrieval queries and progressively accumulate evidence.

    \item \textbf{WebShaper~\cite{tao2025webshaper}.}
    WebShaper formulates information seeking as an iterative query expansion and refinement process, gradually constructing a structured evidence set for answer synthesis. We directly report the WebShaper results published on the WebWeaver benchmark for comparison.

\end{itemize}

\paragraph{\ding{184} Multi-Agent.}
Multi-Agent baselines decompose the task into multiple interacting agents,
leveraging either centralized coordination or automated agent orchestration strategies.
\begin{itemize}
    \item \textbf{STORM~\cite{shao2024assisting}.}
    STORM generates long-form, Wikipedia-style reports through a multi-perspective pipeline that first asks specialized agents to brainstorm an outline from different viewpoints, then conducts grounded conversations to gather evidence, and finally writes the article section by section. Based on their open-source implementation, we conducted experiments adapting the 'bocha' search engine as the retrieval module.
    \item \textbf{WebWeaver~\cite{li2025webweaver}.}
    WebWeaver employs a planner agent and a writer agent that collaborate over a dynamically evolving outline and an evidence memory bank: the planner interleaves search and outline updates, while the writer composes each section by retrieving only the locally relevant evidence to mitigate long-context drift. Building upon their open-source implementation, we integrated the Bocha search engine and conducted experiments.
    \item \textbf{EDR~\cite{prabhakar2025enterprise}.}
    EDR (Enterprise Deep Research) is a steerable, multi-agent system that coordinates specialized domain agents (e.g., Master Planner, Academic/GitHub/LinkedIn Searchers) and enterprise tools (e.g., NL2SQL). It produces comprehensive research reports through iterative refinement, explicit citation tracking, and a reflection mechanism to address knowledge gaps. In our experiments, we built upon the open-source EDR framework and integrated the Bocha search engine to ensure a fair comparison.
    \item \textbf{StackPlanner~\cite{zhang2026stackplanner}.}
    StackPlanner adopts a stack-based planning controller that pushes and pops sub-goals during exploration, allowing the agent to recursively decompose the research task, dispatch sub-queries to retrieval tools, and assemble the final report from the resolved sub-goals.
\end{itemize}

\section{Implementation Details}
\label{appendix:implementation}

We implement \M~without any additional fine-tuning. All LLM interactions are performed through API calls, using \texttt{Qwen3-32B} or DeepSeek for both outline planning and report generation. We adopt BoCha as the default web search backend, retrieving five documents during initialization and eight documents for each subsequent expansion or update query. The outline optimization process is terminated after 20 iterations.

All retrieved documents are encoded using \texttt{text-embedding-3-small}. The resulting 1,536-dimensional embeddings are used to compute all similarity-based metrics, including relevance, novelty, memory retrieval, and sibling-cohesion scores via cosine similarity. We do not use a local encoder for embedding-based retrieval. Instead, local models are reserved for citation verification. Specifically, the system first employs a \texttt{sentence-transformers} CrossEncoder based on \texttt{nli-deberta-v3-small}; when unavailable, DeepSeek is used as a fallback natural language inference (NLI) verifier.

For all experiments, we set $\lambda_1=\lambda_2=\lambda_3=1/3$. Within the utility formulation, we use $w_{\mathrm{rel}}=0.6$ and $w_{\mathrm{nov}}=0.4$ for retrieval, $w_{\mathrm{coh}}=0.4$, $w_{\mathrm{bal}}=0.2$, and $w_{\mathrm{red}}=0.4$ for structural optimization, and $\alpha=0.4$, $\beta=0.4$, and $\gamma=0.2$ for generation.
Following the official DeepResearch Bench protocol, we use GPT-5.5 for RACE and GPT-5.4-mini for FACT. For DeepResearch Gym, we use GPT-5.4-mini as the LLM judge for all methods.

\begin{table*}[t]
\centering
\small
\begin{tabular}{l l l l}
\toprule
\textbf{Utility} & \textbf{Signal} & \textbf{Estimator} & \textbf{Weight} \\
\midrule
\multirow{2}{*}{$U_{\mathrm{ret}}$}
& $\mathrm{Rel}$ & Cosine similarity with \texttt{text-embedding-3-small} & $0.6$ \\
& $\mathrm{Nov}$ & $1-\max$ cosine similarity to cached evidence & $0.4$ \\
\midrule
\multirow{3}{*}{$U_{\mathrm{str}}$}
& $\mathrm{Coh}$ & LLM judge / sibling semantic consistency & $0.4$ \\
& $\mathrm{Bal}$ & Local depth and branching statistics & $0.2$ \\
& $\mathrm{Red}$ & Sibling similarity and LLM redundancy check & $0.4$ \\
\midrule
\multirow{3}{*}{$U_{\mathrm{gen}}$}
& $\rho_{\mathrm{sup}}$ & CrossEncoder NLI, with DeepSeek fallback & $0.4$ \\
& $\rho_{\mathrm{cov}}$ & LLM judge for intent coverage & $0.4$ \\
& $\rho_{\mathrm{red}}$ & Semantic similarity and LLM redundancy check & $0.2$ \\
\bottomrule
\end{tabular}
\caption{Implementation details of utility signals in \M. All component scores are normalized to $[0,1]$. Larger values indicate stronger presence of the corresponding signal; redundancy signals are used as penalty terms in the utility function.}
\label{tab:factstruct-utility-impl}
\end{table*}

\section{Additional Experiments}
\label{sec:additional-experiments}

\paragraph{\ding{182} Clinical Report Case Study}

\begin{figure*}[htbp]
    \centering
    \includegraphics[width=1\linewidth]{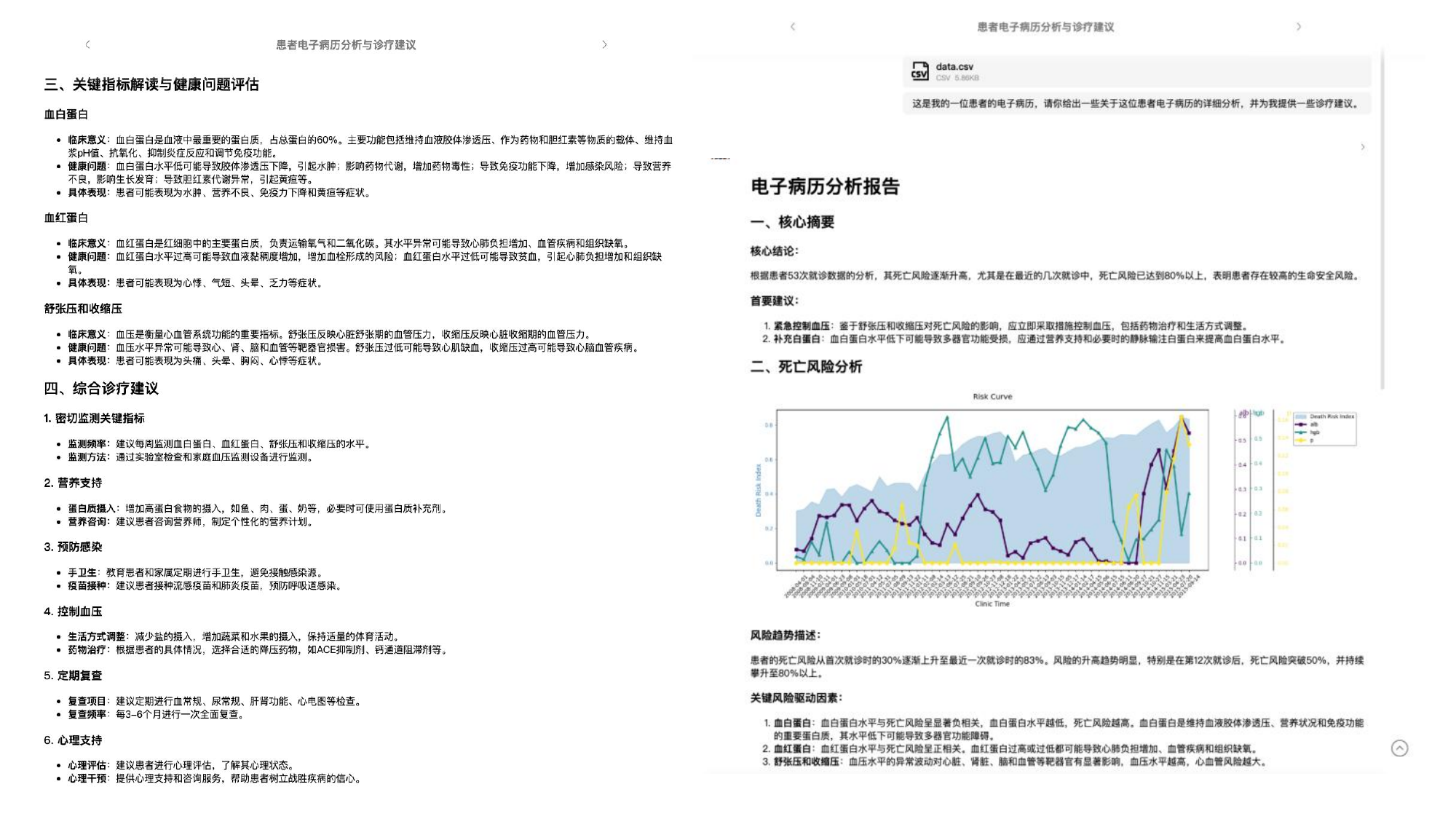}
    \caption{Example clinical report generated by \M (in Chinese).}
    \label{fig:clinical_case}
\end{figure*}

 LLMs have also demonstrated strong potential in healthcare applications~\cite{xu2025dearllm,jiang2024tc,fang2026graphwalker}. To examine the domain generality of \M, we apply it to a clinical reporting task that requires synthesizing evidence from multiple medical sources as shown in Figure~\ref{fig:clinical_case}.

\paragraph{\ding{183} Inference-Cost Comparison.}
Beyond quality, we also report inference cost as a standalone efficiency check. Table~\ref{tab:cost-comparison} shows the average
per-query inference cost. \M~consumes 26.3k tokens, issues 8.2 search calls, and finishes in 116.7\,s, using 29\% fewer tokens and 49\% fewer search calls than WebWeaver (37.2k / 15.9), and running faster than EDR (135.5\,s) at comparable token cost. The utility-guided action selection concentrates retrieval on high-utility nodes, so the RACE and FACT gains in Sections~\ding{182} and \ding{183} do not come at the price of runaway inference cost.

\begin{table}[t]
\centering
\fontsize{9pt}{9pt}\selectfont
\setlength{\tabcolsep}{4pt}
\renewcommand{\arraystretch}{1.1}

\begin{tabular}{l|ccc|cc}
\toprule
\rowcolor{gray!10}
\textbf{Method} & \textbf{Input} & \textbf{Output} & \textbf{Total} & \textbf{\#Search} & \textbf{Latency} \\
\midrule

\textbf{Ours} & \textbf{19.2k} & 7.1k & \textbf{26.3k} & \textbf{8.2} & \textbf{116.7\,s} \\

\midrule

WebWeaver              & 27.9k & 9.3k & 37.2k & 15.9 & 148.5\,s \\
EDR$^{\dagger}$        & 26.0k &\textbf{ 3.7k} & 29.8k & 9.1  & 135.5\,s \\

\bottomrule
\end{tabular}
\caption{Inference cost comparison on DeepResearch Bench.}
\label{tab:cost-comparison}
\end{table}

\paragraph{\ding{184} Report Length Distribution}

\begin{figure}[t]
\centering
\includegraphics[width=\linewidth]{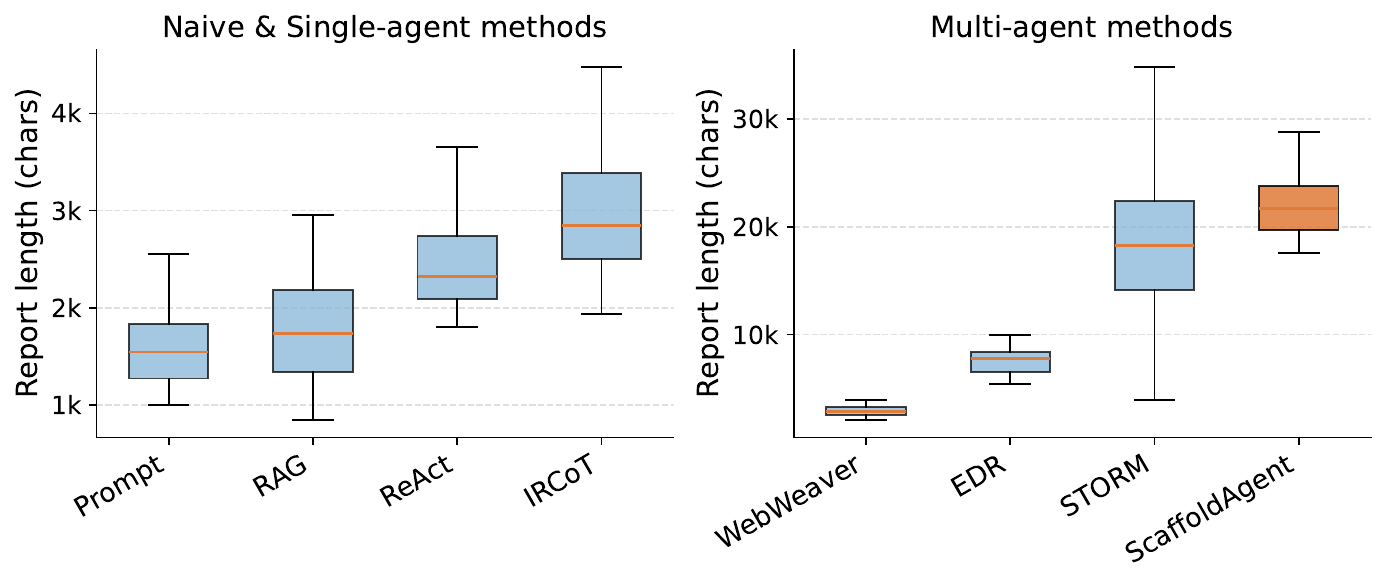}
\caption{Distribution of output report lengths across methods.}
\label{fig:report_length}
\end{figure}

Figure~\ref{fig:report_length} compares the distribution of generated report lengths across methods. Naive and single-agent methods mostly produce short reports below 4K characters, whereas multi-agent methods show substantially longer outputs. Among them, \M~has the highest median length, around 22K characters, and a relatively compact interquartile range compared with STORM, whose outputs vary widely from short reports to over 30K characters. EDR is more concentrated around 6--9K characters, and WebWeaver remains close to the short-report regime. These patterns suggest that \M's optimized outline scaffold supports more extensive and stable synthesis, while the main quality metrics confirm that the longer outputs are accompanied by stronger depth, breadth, and grounding.

\paragraph{\ding{185} Backbone and Search-Backend Generalization.}
Table~\ref{tab:generalization} evaluates whether \M transfers to stronger proprietary backbones and alternative retrieval systems. DeepSeek-V4-Pro and GPT-5.4-mini use the full evaluation setting, whereas
GPT-5.4 is evaluated as a shared 20-query pilot because of its higher API cost. For retrieval generalization, we hold the Qwen3-32B backbone
and all non-retrieval configurations fixed and replace the original web API with Bing Search or a Wikipedia backend constructed from the November 2023 dump.

\begin{table*}[t]
\centering
\fontsize{8.5pt}{9.5pt}\selectfont
\setlength{\tabcolsep}{5pt}
\renewcommand{\arraystretch}{1.08}
\begin{tabular}{ll|l|ccc}
\toprule
\rowcolor{gray!10}
\textbf{Factor} & \textbf{Setting} & \textbf{Method} &
\textbf{RACE} & \textbf{Eff.c.} & \textbf{C.acc.} \\
\midrule
\multirow{6}{*}{Backbone}
& \multirow{2}{*}{DeepSeek-V4-Pro}
& StackPlanner & 49.83 & 38.58 & 59.82 \\
& & \textbf{\M} & \textbf{51.61} & \textbf{54.32} & \textbf{69.21} \\
\cmidrule{2-6}
& \multirow{2}{*}{GPT-5.4-mini}
& StackPlanner & 50.66 & 42.84 & 64.58 \\
& & \textbf{\M} & \textbf{51.35} & \textbf{57.60} & \textbf{72.43} \\
\cmidrule{2-6}
& \multirow{2}{*}{GPT-5.4 (20-query pilot)}
& StackPlanner & 52.18 & 61.63 & 76.90 \\
& & \textbf{\M} & \textbf{53.82} & \textbf{84.27} & \textbf{82.35} \\
\midrule
\multirow{6}{*}{Retrieval}
& \multirow{2}{*}{Original Web API}
& StackPlanner & 41.55 &  9.63 & 26.50 \\
& & \textbf{\M} & \textbf{44.70} & \textbf{30.42} & \textbf{54.32} \\
\cmidrule{2-6}
& \multirow{2}{*}{Bing Search API}
& StackPlanner & 43.84 & 17.62 & 32.14 \\
& & \textbf{\M} & \textbf{46.12} & \textbf{40.85} & \textbf{63.37} \\
\cmidrule{2-6}
& \multirow{2}{*}{Wikipedia Retrieval}
& StackPlanner & 40.35 & 16.70 & 29.85 \\
& & \textbf{\M} & \textbf{42.61} & \textbf{36.86} & \textbf{58.21} \\
\bottomrule
\end{tabular}
\caption{Generalization across stronger backbone models and search
backends. DeepSeek-V4-Pro and GPT-5.4-mini use the full evaluation set;
GPT-5.4 uses a shared 20-query pilot subset. Retrieval-backend results
use Qwen3-32B with all non-retrieval configurations fixed.}
\label{tab:generalization}
\end{table*}

\M consistently outperforms StackPlanner under every tested backbone and retrieval backend. Bing Search produces the strongest results among the retrieval systems, while Wikipedia's narrower coverage lowers RACE but still preserves substantial gains in effective citation coverage and citation accuracy. The consistent method-level advantage indicates that \M is not tied to a particular generator or search API.

\paragraph{\ding{186} Utility-Weight Sensitivity.}
Tables~\ref{tab:main-weight-sensitivity} and \ref{tab:generation-weight-sensitivity} vary the top-level utility weights and the sub-weights of $U_{\mathrm{gen}}$ on Qwen3-32B, with all other settings fixed. Top-level perturbations change RACE by at most $1.10$ points; emphasizing retrieval improves citation metrics, whereas emphasizing generation gives the best RACE. Within $U_{\mathrm{gen}}$, increasing support raises C.acc. from 54.32 to 62.92 but reduces RACE and Eff.c., exposing a precision--coverage
trade-off. The default is therefore a balanced operating point rather
than a narrowly tuned optimum.

\begin{table}[t]
\centering
\fontsize{8pt}{9pt}\selectfont
\setlength{\tabcolsep}{2.6pt}
\renewcommand{\arraystretch}{1.05}
\begin{tabular}{l|c|ccc}
\toprule
\rowcolor{gray!10}
\textbf{Bias} & \textbf{Weights} &
\textbf{RACE} & \textbf{Eff.c.} & \textbf{C.acc.} \\
\midrule
Balanced   & $(1/3,1/3,1/3)$ & 44.70 & 30.42 & 54.32 \\
Retrieval  & $(.50,.25,.25)$ & 43.81 & \textbf{32.54} & \textbf{58.07} \\
Structure  & $(.25,.50,.25)$ & 44.03 & 27.86 & 52.41 \\
Generation & $(.25,.25,.50)$ & \textbf{44.91} & 30.68 & 56.36 \\
\bottomrule
\end{tabular}
\caption{Sensitivity to top-level utility weights
$(\lambda_{\mathrm{ret}},\lambda_{\mathrm{str}},\lambda_{\mathrm{gen}})$
on Qwen3-32B. Bold marks the best score in each column.}
\label{tab:main-weight-sensitivity}
\end{table}

\begin{table}[t]
\centering
\fontsize{8.2pt}{9.2pt}\selectfont
\setlength{\tabcolsep}{4pt}
\renewcommand{\arraystretch}{1.05}
\begin{tabular}{c|ccc}
\toprule
\rowcolor{gray!10}
\textbf{$U_{\mathrm{gen}}$ Weights} &
\textbf{RACE} & \textbf{Eff.c.} & \textbf{C.acc.} \\
\midrule
$(0.40,0.40,0.20)$ & \textbf{44.70} & \textbf{30.42} & 54.32 \\
$(0.50,0.25,0.25)$ & 44.66 & 30.18 & 55.28 \\
$(0.70,0.15,0.15)$ & 43.62 & 27.85 & 60.74 \\
$(0.90,0.05,0.05)$ & 41.31 & 23.46 & \textbf{62.92} \\
\bottomrule
\end{tabular}
\caption{Sensitivity to $U_{\mathrm{gen}}$ weights (support, coverage,
and redundancy) on Qwen3-32B. Bold marks the best score in each column.}
\label{tab:generation-weight-sensitivity}
\end{table}

\paragraph{\ding{187} Repeated-Evaluation Stability.}
\label{app:repeated-evaluation-stability}
To assess robustness to stochasticity in LLM-based evaluation, we apply the same RACE evaluator three times to the same fixed Qwen3-32B-generated reports, without changing the evaluator configuration. Table~\ref{tab:evaluator-stability} reports the mean and within-report
standard deviation across runs. Across three repeated evaluations, \M retains higher mean scores in Overall, Comprehensiveness, Insight, and Instruction Following, indicating that the main performance trend is stable under repeated judging.

\begin{table*}[t]
\centering
\fontsize{8.2pt}{9.2pt}\selectfont
\setlength{\tabcolsep}{3.2pt}
\renewcommand{\arraystretch}{1.08}
\begin{tabular}{l|ccccc}
\toprule
\rowcolor{gray!10}
\textbf{Method} & \textbf{Overall} & \textbf{Comp.} &
\textbf{Insight} & \textbf{Inst.} & \textbf{Read.} \\
\midrule
StackPlanner & $41.15{\pm}1.74$ & $39.85{\pm}2.43$ &
$39.23{\pm}2.48$ & $44.11{\pm}2.56$ & $44.29{\pm}1.51$ \\
\M & $44.79{\pm}1.16$ & $43.55{\pm}1.40$ &
$45.35{\pm}1.49$ & $46.53{\pm}1.34$ & $42.83{\pm}1.50$ \\
\bottomrule
\end{tabular}
\caption{Stability across three repeated RACE evaluations of the same
fixed Qwen3-32B-generated reports.}
\label{tab:evaluator-stability}
\end{table*}

\section{Computational Resources and Software Environment}
Experiments were conducted using API-based inference. The Qwen-series backbone was deployed on a single-machine multi-GPU server equipped with 2 NVIDIA A800 GPUs with 80GB memory each, dual Intel Xeon Platinum 8358 CPUs with 64 physical cores and 128 threads in total, and 1024GB of DDR4 memory. The server ran Ubuntu 22.04.5 LTS with Docker, and the inference service was built with vLLM 0.13.0 under FP16 precision. The software environment used Python 3.13.12 and HuggingFace Transformers 4.57.3. On the client side, our agent framework was implemented with Python 3.11, LiteLLM 1.63.11, OpenAI 2.8.1, MCP 1.6.0, LangChain 0.3.25, and LangGraph 0.4.10 for API calls, tool interaction, and agent orchestration.

\section{The Use of Large Language Models}

In this work, Large Language Models (LLMs) were used for language polishing, figure design assistance, and coding support. Specifically, LLMs were used to refine the clarity, grammar, and readability of the manuscript, improve stylistic quality, and suggest alternative phrasings to reduce redundant or repetitive expressions. LLMs were also used to provide inspiration for figure design, visualization layouts, and graphical presentation, with final figures created and curated by the authors. In addition, LLMs assisted with code development by suggesting code snippets and troubleshooting strategies. All content generated by LLMs was carefully reviewed, verified, and revised by the authors before inclusion. The research design, methodology, critical analyses, and all final decisions were independently conducted by the authors. LLMs were not involved in generating the core research ideas, scientific contributions, or conclusions of this work.

\newpage

\section{Prompt}
\label{sec:prompts}
In this section, we provide a detailed introduction to the prompts used in
the \M framework. 

\begin{tcolorbox}
[colback=lightgray!20,colframe=darkgray!80,title=Initial Outline Generation Prompt]
\label{tab:factstruct_initial_outline_prompt}

You are a research assistant. Your task is to generate an initial research outline scaffold for iterative refinement, rather than a complete final outline.

\textbf{User Query:}

\texttt{\{query\}}

\textbf{Initial Evidence:}

\texttt{\{initial\_documents\}}

\textbf{User Constraints:}

\texttt{\{user\_constraints\}}
 
Before generating the outline, internally identify the required analysis dimensions, target entities, comparison requirements, and expected output type in the user query. The outline must cover all explicit requirements in the query and avoid irrelevant background sections.

Generate a minimal usable outline scaffold. Prefer 3--5 top-level sections. Add second-level sections only when necessary. Do not over-expand the outline at initialization time.

\textbf{\textit{Response Format:}}

Return only a JSON object in the same language as the user query:

\texttt{\{}
\texttt{"title": "root title",}
\texttt{"children": [\{"title": "section title", "children": []\}]}
\texttt{\}}
\end{tcolorbox}

\begin{tcolorbox}
[colback=lightgray!20,colframe=darkgray!80,title=Outline Operation Decision Prompt]
\label{tab:factstruct_outline_decision_prompt_aligned}

You are the Outline Agent in ScaffoldAgent. At each optimization step, your task is to decide how the evolving outline tree should be updated by selecting the appropriate structural operation for the chosen outline node.

\textbf{User Query:}

\texttt{\{user\_query\}}

\textbf{User Constraints:}

\texttt{\{user\_constraints\}}

\textbf{Current Outline Tree:}

\texttt{\{outline\_tree\}}

\textbf{Selected Target Node:}

\texttt{\{selected\_target\_node\}}

\textbf{Node Utility Statistics:}

\texttt{\{node\_utility\_statistics\}}

If \texttt{selected\_target\_node} is provided, keep it as the target node and choose only the operation. If it is not provided, select a target node by considering historical utility feedback, visit counts, evidence support, user constraints, and whether the node remains under-developed or unstable. Then choose exactly one operation from:

\texttt{Expansion}, \texttt{Contraction}, \texttt{Revision}.

Use \texttt{Expansion} when the selected node is too broad or lacks sufficient topical granularity for focused retrieval and writing. Use \texttt{Contraction} when sibling nodes are redundant, semantically overlapping, or overly fragmented. Use \texttt{Revision} when the selected node is weakly supported, outdated, or misaligned with its accumulated evidence, while the tree structure should remain unchanged.

\textbf{\textit{Response Format:}}

Return only one JSON object:

\texttt{\{}
\texttt{"target\_node": "selected or newly chosen node id/title",}
\texttt{"operation": "Expansion | Contraction | Revision",}
\texttt{"reasoning": "brief reason grounded in utility statistics and diagnostics",}
\texttt{\}}
\end{tcolorbox}

\begin{tcolorbox}
[colback=lightgray!20,colframe=darkgray!80,title=Expansion Prompt]
\label{tab:factstruct_expansion_prompt}

You are the Expansion operator in ScaffoldAgent. New evidence has been retrieved for a selected outline node. Your task is to locally expand this target node into a small set of evidence-supported child nodes.

\textbf{Original User Query:}

\texttt{\{query\}}

\textbf{User Constraints:}

\texttt{\{user\_constraints\}}

\textbf{Current Outline Tree:}

\texttt{\{current\_outline\}}

\textbf{Target Node:}

\texttt{\{target\_node\}}

\textbf{Retrieved Evidence:}

\texttt{\{retrieved\_documents\}}

Perform an independent local optimization for the target node. Expand only the target node into a small set of parallel child nodes, typically 2--4. Each new child node should represent a distinct subtopic, be specific enough for focused retrieval and writing, and be directly supported by the retrieved evidence. Do not modify unrelated nodes, parent nodes, or sibling nodes. The refined subtree must remain aligned with the original user query and user constraints.

\textbf{\textit{Response Format:}}

Return only the JSON subtree for the target node, in the same language as the user query:

\texttt{\{}
\texttt{"title": "target node title",}
\texttt{"children": [}
\texttt{\{"title": "expanded child title", "children": []\}}
\texttt{]}
\texttt{\}}
\end{tcolorbox}

\begin{tcolorbox}
[colback=lightgray!20,colframe=darkgray!80,title=Contraction Prompt]
\label{tab:factstruct_compression_prompt}

You are the Contraction operator in ScaffoldAgent. Your task is to locally consolidate sibling nodes under a selected parent node when they are semantically overlapping, overly fragmented, redundant, or supported by similar evidence.

\textbf{Original User Query:}

\texttt{\{query\}}

\textbf{User Constraints:}

\texttt{\{user\_constraints\}}

\textbf{Current Outline Tree:}

\texttt{\{current\_outline\}}

\textbf{Target Parent Node:}

\texttt{\{parent\_node\}}

\textbf{Sibling Nodes to Compress:}

\texttt{\{child\_nodes\}}

\textbf{Existing Evidence Summaries:}

\texttt{\{evidence\_summaries\}}

Perform an independent local contraction under the target parent. Preserve the parent node title, the parent node's responsibility, and the global outline direction. Only modify the children of the target parent; do not change unrelated nodes, ancestors, or sibling groups outside this parent. Merge overlapping, redundant, or overly fragmented children into representative child nodes. The contracted children must jointly preserve the key information of the original children, remain grounded in the existing evidence, and stay aligned with the original user query and user constraints.

The number of children after contraction should be smaller than the original number whenever consolidation is justified, while avoiding over-compression. As an implementation guideline, 3 children are typically contracted into 2; 4 children into 2 or 3; and 5 or more children into approximately half of the original number. Keep at least two children whenever possible.

\textbf{\textit{Response Format:}}

Return only the JSON subtree for the target parent, in the same language as the user query:

\texttt{\{}
\texttt{"title": "parent node title",}
\texttt{"children": [}
\texttt{\{"title": "contracted child title", "children": []\}}
\texttt{]}
\texttt{\}}
\end{tcolorbox}

\begin{tcolorbox}
[colback=lightgray!20,colframe=darkgray!80,title=Revision Prompt]
\label{tab:factstruct_revision_prompt}

You are the Revision operator in ScaffoldAgent. Your task is to locally revise the intent or titles of a selected node's local child nodes so that they are better aligned with the available evidence, diagnostics, and user constraints. This task is not Expansion or Contraction. The outline structure must remain unchanged.

\textbf{Original User Query:}

\texttt{\{query\}}

\textbf{User Constraints:}

\texttt{\{user\_constraints\}}

\textbf{Current Outline Tree:}

\texttt{\{current\_outline\}}

\textbf{Selected Node:}

\texttt{\{selected\_node\}}

\textbf{Local Child Nodes:}

\texttt{\{child\_nodes\}}

\textbf{Evidence and Diagnostics:}

\texttt{\{evidence\_and\_diagnostics\}}

Revise the selected node's local child titles to make them more specific, informative, and better grounded in the available evidence. Do not add child nodes. Do not remove child nodes. Do not change the hierarchy. Do not change the selected node title. Revise at least one local child title when evidence misalignment, weak support, or outdated wording is identified. The revised child titles must preserve the original responsibility of the selected node and remain aligned with the original user query and user constraints.

\textbf{\textit{Response Format:}}

Return only the JSON subtree for the selected node, in the same language as the user query:

\texttt{\{}
\texttt{"title": "selected node title",}
\texttt{"children": [}
\texttt{\{"title": "updated child title", "children": []\}}
\texttt{]}
\texttt{\}}
\end{tcolorbox}

\end{document}